\documentclass{article}
\usepackage{spconf,amsmath,graphicx,multirow}
\usepackage{booktabs}
\usepackage{natbib}
\setcitestyle{numbers,square}

\title{Prompt-based Graph Model for Joint Liberal Event Extraction and Event Schema Induction}
\name{Haochen Li, Di Geng, Tong Mo, Siyuan Lu, Weiping Li\thanks{This work is supported by the National Key Research and Development Program of China (No. 2020YFC0833300). Haochen Li and Di Geng contributed equally to this work.}}
\address{Peking University\\
        School of Software and Microelectronics\\
        Beijing, China}
\begin{document}
%
\maketitle
\begin{abstract}
Events are essential components of speech and texts, describing the changes in the state of entities. The event extraction task aims to identify and classify events and find their participants according to event schemas. 
Manually predefined event schemas have limited coverage and are hard to migrate across domains. Therefore, the researchers propose Liberal Event Extraction (LEE), which aims to extract events and discover event schemas simultaneously.
However, existing LEE models rely heavily on external language knowledge bases and require the manual development of numerous rules for noise removal and knowledge alignment, which is complex and laborious.
To this end, we propose a \textbf{P}rompt-based \textbf{G}raph Model for \textbf{L}iberal \textbf{E}vent \textbf{E}xtraction \textbf{(PGLEE)}.
Specifically, we use a prompt-based model to obtain candidate triggers and arguments, and then build heterogeneous event graphs to encode the structures within and between events.
Experimental results prove that our approach achieves excellent performance with or without predefined event schemas, while the automatically detected event schemas are proven high quality.

\end{abstract}
\section{Introduction}
\label{sec:intro}

\begin{figure}[htb]

  \centering
  \centerline{\includegraphics[width=7cm]{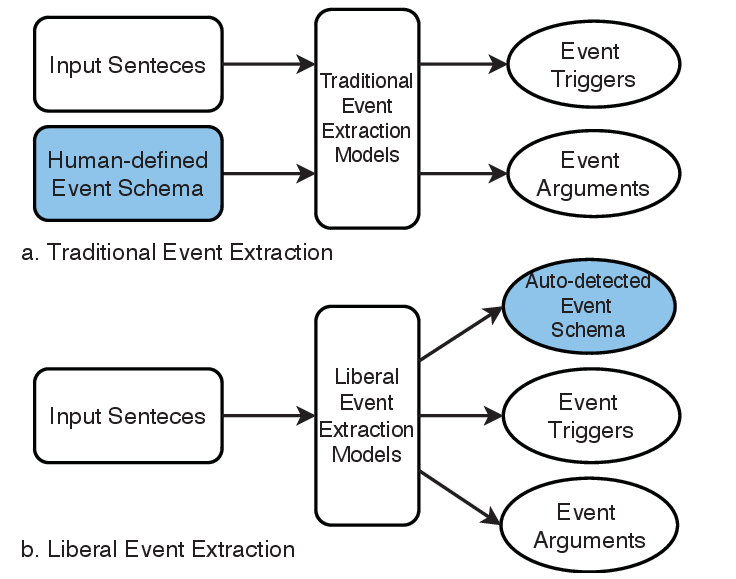}}
  \caption{The target of Liberal Event Extraction.}
\end{figure}

Events describe state changes of participating entities. Discovering and tracking events in conversations and texts is helpful in downstream tasks such as intent tracking and natural language inference. Event triggers are words or phrases most representative of an event, while event arguments are entities participating in the event. The event extraction task aims to identify and classify event triggers and arguments.

\begin{figure*}[htb]

  \centering
  \centerline{\includegraphics[width=\textwidth]{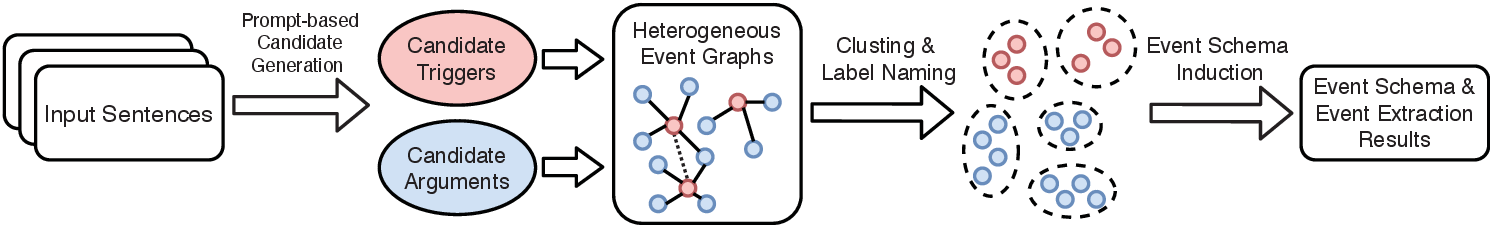}}
  \caption{The overall structure of PGLEE.}
\end{figure*}

However, most existing event extraction approaches require manually predefined event schemas as antecedent information (Fig.1a). Manual templates have the following drawbacks: A. Low resource, the number and coverage of manually defined event schema by domain experts are limited; B. Hard to migrate, the event schemas need to be redefined when the domain change. Therefore, only relying on manually defined event schemas limits the wide application of event extraction methods.

Huang et al.\cite{huang_liberal_2016} first propose Liberal Event Extraction (LEE) to address the above problems, which automatically discovers event schemas and extracts events simultaneously. Nguyen et al.\cite{nguyen_generative_2015} and Sha et al.\cite{sha_joint_2016} use entity linking and entity disambiguation for event schema induction, respectively. However, these efforts have the following shortcomings:
\textbf{a.} They rely heavily on semantic analysis tools and external knowledge bases and require manual rules to remove noise and construct alignment mappings between multiple language resources.
\textbf{b.} They only consider the effect of arguments on trigger classification, ignoring the reverse impact and the inter structure of events.
\textbf{c.} Modules in the model are connected in the form of pipelines which can lead error propagation. 

This paper focuses on the Liberal Event Extraction (LEE) task (Fig.1b). We propose a Prompt-based Graph Model for Liberal Event Extraction (PGLEE) to implement end-to-end LEE task. First, we use a prompt-based method to generate candidate triggers and arguments. Further, we construct heterogeneous event graphs to enhance the information interactions within and between the events, and then obtain their semantic embeddings through graph representation learning algorithms. Finally, we use clustering algorithms and label naming to discover event schemas and extract events simultaneously. Our contributions can be summarized as follows:

\begin{itemize}
	\item We focus on the LEE task and propose a novel prompt-based graph model PGLEE, achieving end-to-end event extraction and event schema induction.
	\item A prompt-based candidate generation model is proposed for event triggers and arguments, eliminating external knowledge bases and manual rules. 
	\item Heterogeneous event graphs are constructed to enhance the feature interactions. Event extraction and new event schema induction are performed simultaneously by clustering and label naming.
 	\item Experimental results prove the effectiveness of  our approach, while the automatically detected event schemas are proven high quality.
\end{itemize}

\section{Approach}

\subsection{Event Candidates Generation}
Given the input sentences, we aim to obtain candidate events (Event $t$ has arguments: $a_1, ... , a_n$) . Previous approaches rely heavily on semantic parsing tools (e.g., Stanford NLP, AMR) and external knowledge bases (e.g., WordNet, VerbNet). These approaches require multiple manual rules to align the semantic analysis results with the knowledge base's resources, which is cumbersome and difficult to migrate to new domains. We use an intuitive prompt-based model to generate output directly from the target sentences without using additional knowledge bases.

\begin{table}[]
\small
\caption{An example of the prompt template.}
\begin{tabular}{c|l}
\toprule
\textbf{x (input)}         & \begin{tabular}[c]{@{}l@{}}The threat posed by the \textit{Iraqi dictator} justif-\\ies  a \textbf{war}, which is sure to \textbf{kill} thousands of\\innocent \textit{children} and \textit{women}.\end{tabular} \\ \hline
\textbf{prompt(x)} & \begin{tabular}[c]{@{}l@{}}threat posed Iraqi dictator justifies war kill\\children women. \end{tabular}                                                            \\ \hline
\textbf{y}         & \begin{tabular}[c]{@{}l@{}} Event \textbf{war} has arguments: \textit{Iraqi dictator}; \\ Event \textbf{kill} has arguments: \textit{children, women}.\end{tabular}                                   \\
\bottomrule
\end{tabular}

\end{table}

Specifically, for an input $x$, we construct the following prompt template using the prefix-prompt strategy \cite{li_prefix-tuning_2021}.

\begin{equation}
	Template: [x], prompt(x), [y]
\end{equation}
Where $prompt(x)$ is the input-related prompt associated with $x$ and $y$ is the generated result. As in the example in Table 1, the sentence has two events, “war” and “kill”, and the second event contains two arguments, “children” and “women”. We add the nouns and verbs (event triggers are primarily nouns and verbs) and entities (event arguments are entities) in the sentence to $prompt(x)$. The generated target $y$ is a semi-structured text containing candidate event triggers and arguments. Following the previous work\cite{li_kipt_2022}, we also add 20 soft tokens (virtual tokens sharing the same dimension as actual words but without real meanings ) to enhance the performance of the prompt model.

\subsection{Event Graph Construction}

For each input sentence $x$, candidate events can be generated by the method in 2.1: Event $t_i$ has arguments: $a_{i1}, ... , a_{in}$.
Then, the heterogeneous event graph $G_{event}$ is constructed using the results above (as shown in the middle part of Fig.2). there are two kinds of nodes in $G_{event}$: candidate triggers (red nodes in Fig.2) and arguments (blue nodes in Fig.2). There are two kinds of relational edges in $G_{event}$: trigger-argument edge inside an event  (solid lines in Fig.2) and trigger-trigger edge (dashed line in Fig.2) between events.

We use a similar graph attention networks\cite{velickovic_graph_2022} to enhance semantic embeddings of candidate triggers and arguments, and capture feature interactions within and between events. Specifically, if the node $i$ in $G_{event}$ has a neighbor node $j$. First, we perform feature transformation on the word embeddings of nodes $i$ and $j$ and then:
\begin{equation}
	e_{ij}=< W^{i}h_{i}, W^{j}h_{j} > ,j\in  N_{i}
\end{equation}
\begin{equation}
W^{d}=\left\{
\begin{aligned}
W^{trig}, &\; \text{if node }d\in Trigs \\
W^{arg},  &\; \text{if node }d\in Args
\end{aligned}
\right.
,d \in N_i\; or\; d=i
\end{equation}	
Where $W^{trig}$ and $W^{arg}$ are transformation matrices for candidate event triggers and arguments, $h_{i}$ stands for the original word embedding vector of node $i$, $< \cdot , \cdot > $ stands for the inner product operation, and $N_i$ is the set of all neighbors of node $i$. 
Then we normalize the attention coefficients between node $i$ and all neighbors.
\begin{equation}
	\alpha_{ij} =\frac{exp\left ( LeakyReLU\left ( e_{ij}  \right )  \right ) }{ {\textstyle \sum_{k\in N_{i}}^{}}exp\left ( LeakyReLU\left ( e_{ik}  \right )  \right )  }
\end{equation}
$LeakyReLU$ is the activation function. After normalization, we can obtain the new representation of each node.
\begin{equation}
	h_{i}^{'} =\sigma \left (  {\textstyle \sum_{j\in N_{i} }^{}} \alpha _{ij} W^{j}h_{j}  \right )
\end{equation}
The above equation is extended by multi-head attention \cite{vaswani_attention_2017}, and the results of multiple heads are averaged.
\begin{equation}
	h_{i}^{'} =\frac{1}{K}\sum_{l=1}^{K}\sigma \left (  {\textstyle \sum_{j\in N_{i} }^{}} \alpha _{ij} W^{j}_{l}h_{j}  \right ) 
\end{equation}
Where $K$ stands for the number of attention heads, and $W^{j}_{l}$ stands for the transformation matrices of the $l_{th}$ head. Finally, we can obtain feature-interacting semantic representations of all candidate elements.

\subsection{Clustering and Event Schema Induction}
After encoding in 2.2, nodes of the same type are close in the feature space \cite{zhao_graphsmote_2021}. Thus, we use the Mini-batch K-means clustering\cite{peng_clustering_2018} algorithm to obtain candidate trigger labels and argument role types. We set a fixed number of iterations in each training epoch and finally generate $k^{trig}$ clusters of event types and $k^{arg}$ clusters of argument role types, respectively. For node $i$ (assuming $i$ is a candidate trigger), we use its distance to each cluster to calculate the probability of belonging to the cluster.
\begin{equation}
	p(i\in cluster_t) = \frac{dist(i, c_t)^{-1}}{\sum_{j=1}^{k^{trig}}dist(i, c_j)^{-1}}
\end{equation}
Where $dist(\cdot ,\cdot )$ stands for the Euclidean distance and $c_j$ stands for the center of the $j$th cluster. In choosing $k^{trig}$ and $k^{arg}$ fetches, we have two ways: a. picking the same number of labels as in the public dataset; b. freely exploring the number of event types and the number of argument role types. The first approach aims to test our model's performance in supervised learning and we map the clustering results to the existing labels. And the second approach aims to discover new event and argument types, and we determine the optimal choice by the Silhouette coefficient\cite{hruschka_feature_2005}.
For the label naming of the candidate triggers, we select the text of the node closest to the cluster center as the label name of that cluster. For the candidate arguments, we do label naming manually because of the immense diversity of entity texts. To achieve automatic event schema induction, we look for all argument roles with edges in the event graphs for each event type. If the edge has an attention coefficient over the threshold $\theta$, we add the corresponding argument to the event schema.
\section{Experiments}
\subsection{Experiment Setup}

\begin{table*}[]
\centering
\small
\caption{Experimental results on TAC-KPB 2017. We compare the performance of the models trained on different numbers of predefined event schemas. The average F1 scores of 10 experiments are used for each model.}
\begin{tabular}{l|cccc|cccc|cccc}
\toprule
\multicolumn{1}{c|}{\textbf{Models}} & \multicolumn{4}{c|}{\textbf{Results with 5 event schemas}} & \multicolumn{4}{c|}{\textbf{Results with 10 event schemas}} & \multicolumn{4}{c}{\textbf{Results with all event schemas}} \\ \hline
\multicolumn{1}{c|}{\textbf{}} & \textbf{Trig-I} & \textbf{Trig-C} & \textbf{Arg-I} & \textbf{Arg-C} & \textbf{Trig-I} & \textbf{Trig-C} & \textbf{Arg-I} & \textbf{Arg-C} & \textbf{Trig-I} & \textbf{Trig-C} & \textbf{Arg-I} & \textbf{Arg-C} \\ \hline
Joint3EE (2019) & 40.8 & 33.4 & 24.6 & 19.2 & 53.6 & 46.1 & 37.7 & 31.9 & \textbf{70.5} & \textbf{62.1} & 52.9 & 46.3 \\
OneIE (2020) & 39.5 & 31.6 & 25.8 & 20.9 & 52.1 & 44.4 & 38.5 & 33.0 & 68.4 & 57.0 & 50.1 & 46.5 \\ \hline
PoKE (2021) & 41.1 & 33.2 & 28.9 & 25.1 & 58.0 & 49.6 & 42.8 & 39.0 & 66.3 & 57.9 & 51.1 & \textbf{47.4} \\
KiPT (2022) & 42.3 & 36.6 & - & - & 58.8 & 50.9 & - & - & 67.0 & 58.3 & - & - \\ \hline
Liberal\_EE (2016) & - & 46.4 & - & 25.4 & - & 57.1 & - & 34.6 & - & 61.8 & - & 39.3 \\
SSED (2018) & 48.8 & 39.8 & - & - & \textbf{64.2} & 53.0 & - & - & 68.7 & 57.5 & - & - \\ \hline
\textbf{PGLEE (Ours)} & \textbf{51.0} & \textbf{47.7} & \textbf{43.9} & \textbf{41.5} & 63.3 & \textbf{58.0} & \textbf{53.2} & \textbf{45.8} & 66.6 & 61.3 & \textbf{53.5} & 47.1 \\
\bottomrule
\end{tabular}

\label{tab:main}
\end{table*}

\begin{table*}[]
\centering
\caption{Examples of the new detected event schemas.}
\begin{tabular}{c|c|c}
\toprule
\textbf{Event Type} &
  \textbf{Event Arguments} &
  \textbf{Example Sentence} \\ \hline
Arrive &
  \begin{tabular}[c]{@{}l@{}}Monday(Time),\\  positions(Place)\end{tabular} &
  \begin{tabular}[c]{@{}l@{}}\textbf{S1}: Palestinian security forces \textbf{returned} \underline{Monday} to the \underline{positions} they held in the \\ Gaza Strip before the outbreak of the 33-month Palestinian uprising as Israel re- \\ moved all checkpoints in the coastal territory, a Palestinian security source said.
  \end{tabular} \\ \hline
  
  
Collide &
  \begin{tabular}[c]{@{}l@{}}Fehmi Husrev Kutlu(Agent),\\  Ozkan(Target)\end{tabular} &
  \begin{tabular}[c]{@{}l@{}}\textbf{S2}: A furious Justice party member, \underline{Fehmi Husrev Kutlu}, \textbf{rushed} toward \underline{Ozkan},  \\ bumping into him and sending his eyeglasses flying, Anatolia reported.\end{tabular} \\ 
  \bottomrule
\end{tabular}

\label{tab:my-table2}
\end{table*}

Our work is evaluated on the widely used event extraction dataset TAC-KPB 2017 (Event Nugget data of TAC 2017)\footnote{https://catalog.ldc.upenn.edu/LDC2020T13}. The dataset mainly comprises documents of speeches, interviews, news dialogues, etc., and predefined 38 event schemas with 24 argument types. We follow the previous dataset splits (\cite{li_document-level_2021}\cite{lu_text2event_2021}) using 8,026/683/572 text samples as training, test and development sets. \\
\textbf{Settings.} In this paper, T5 \cite{yang_exploring_2019} is utilized as the pre-trained language model. Our model use AdamW as the optimizer for 30 training epochs with a learning rate of 1e-4 and a weight decay of 1e-5 for T5, and a learning rate of 1e-3 and a weight decay of 1e-4 for other parameters. The batch size is set to 16 and the hyper-parameter $\theta$ is set to 0.3. \\
\textbf{Metrics.} The Event Extraction task consists of four subtasks: Trigger Identification (Trig-I), Trigger Classification (Trig-C), Argument Identification (Arg-I), and Argument Classification (Arg-C). We use micro-averaged F1 score (F1) in all the following evaluations. \\
\textbf{Baselines.} This paper chooses 6 strong baselines for comparison: (1) Two sequence tagging EE models: \textbf{Joint3EE}\cite{nguyen_one_2019} jointly extracts entities, triggers, and arguments based on the shared hidden representations; \textbf{OneIE}\cite{lin_joint_2020} builds an end-to-end information extraction system based on globally optimal event structures; (2) Two prompt-based learning models: \textbf{PoKE}\cite{lin_eliciting_2021} presents joint prompt methods by modeling the interactions between different triggers or arguments; \textbf{KiPT}\cite{li_kipt_2022} uses external knowledge bases to inject event-related knowledge into the prompt templates. (3) Two Liberal Event Extraction models:  \textbf{Liberal\_EE}\cite{huang_liberal_2016} which is mentioned before; \textbf{SSED}\cite{ferguson_semi-supervised_2018} presents self-training event extraction systems using parallel mentions of the same event instance;

Since our PGLEE does not use predefined event templates, the trigger and argument labels automatically inducted may not match the annotations in the dataset. Therefore, in subsection 3.2, we first restrict our model to generate the same labels as the annotated data to compare our model with the baselines. Then in subsection 3.3, we remove the limits and discuss the new event schemas inducted by our model.

\subsection{Event Extraction for Predefined Types}
We restrict the models to generate the same types of triggers and arguments as the annotated data. So we can compare the performances of PGLEE with baselines when extracting predefined or new event schemas. When training, we set different numbers of predefined event schemas and remove the remaining event types' labels. Then we evaluate the models' extraction performances on all event types in the test dataset. The results are shown in Table 2, from which we can find that: \\
\textbf{a. }PGLEE significantly outperforms the baselines with few predefined event schemas. The F1 score of PGLEE on the four tasks exceeds the strongest baseline by 10.3\% (5 event schemas) and 5.3\% (10 event schemas) on average. This validates the ability of our model to liberally extract events and better discover undefined events and their corresponding arguments compared to the current models.\\
\textbf{b. }The performance improvement of PGLEE is even more significant in argument identification and extraction, averaging 15.5\% (5 event schemas) and 8.6\% (10 event schemas). This reflects the effectiveness of the multiple information interaction within and between the events in our heterogeneous event graphs, while the existing models only emphasize the one-side interaction of arguments to triggers. \\
\textbf{c. }When using all predefined event schemas, PGLEE still achieves competitive performance compared to traditional supervised models, falling behind by only 0.8\% (Trig-C) and 0.3\% (Arg-C). This implies that our method also has reliable performance with sufficient labeled data.

\subsection{New Event Schema Induction}
\begin{figure}[htb]
  \centering
  \centerline{\includegraphics[width=7.2cm]{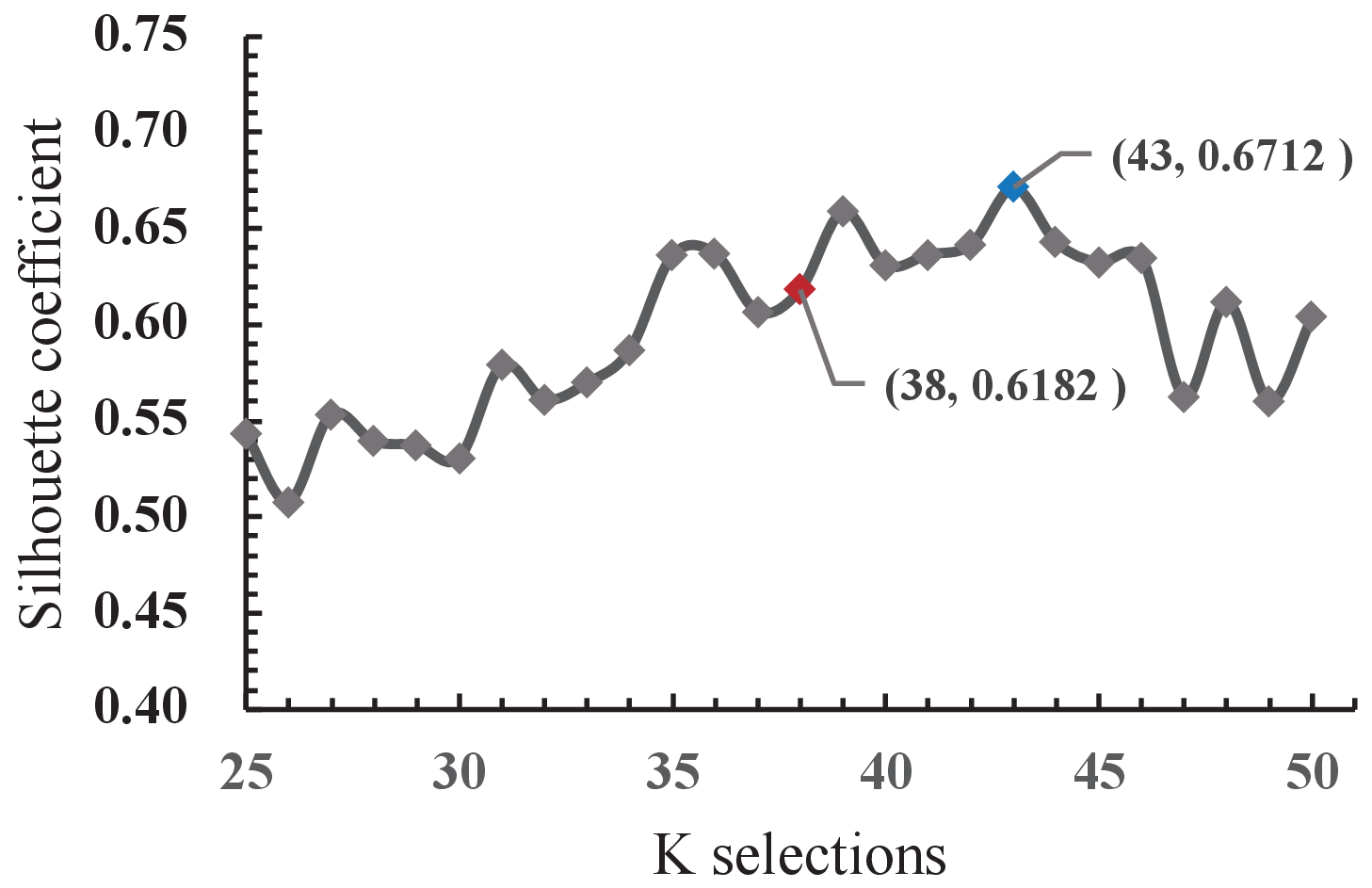}}
  \caption{The Silhouette coefficient of different K selections.}
\end{figure}

We remove the restriction in subsection 3.2 and use the Silhouette coefficient to explore the selection of the number of event types with the best clustering effect. The results under different K selections are shown in Fig. 3: the red point corresponds to the 38 event types given by the TAC-KPB dataset, and the blue point corresponds to the 43 event types which has the optimal clustering performance.

Besides the predefined schemas, PGLEE discovers new event schemas such as \textbf{Arrive} and \textbf{Collide} shown in Table 3. For example, in \textbf{S1}, besides the predefined Attack event, we discover a new type of event Arrive(returned) and corresponding arguments Time(Monday), Place(positions).
\section{Conclusion}
This paper proposes an end-to-end approach called PGLEE to implement joint event extraction and event schema induction. We generate candidate event triggers and arguments directly by a prompt-based learning approach without external knowledge bases. 
Heterogeneous event graphs are constructed to encode semantic interactions within and between the events. We use clustering and label naming to detect events schemas automatically. Experimental results validate the excellent extraction ability of PGLEE on predefined event schemas and the ability to discover new schemas.
We will explore more approaches to liberal event extraction and complex event schema induction in the future.



\bibliographystyle{IEEEbib}
\bibliography{strings,refs}

\begin{thebibliography}{10}

\bibitem{huang_liberal_2016}
Lifu Huang, Taylor Cassidy, Xiaocheng Feng, Heng Ji, Clare~R. Voss, Jiawei Han,
  and Avirup Sil,
\newblock ``Liberal event extraction and event schema induction,''
\newblock in {\em Proceedings of the 54th Annual Meeting of the Association for
  Computational Linguistics (Volume 1: Long Papers)}. pp. 258--268, Association
  for Computational Linguistics.

\bibitem{nguyen_generative_2015}
Kiem-Hieu Nguyen, Xavier Tannier, Olivier Ferret, and Romaric Besan{\c c}on,
\newblock ``Generative event schema induction with entity disambiguation,''
\newblock in {\em Proceedings of the 53rd Annual Meeting of the Association for
  Computational Linguistics and the 7th International Joint Conference on
  Natural Language Processing (Volume 1: Long Papers)}. pp. 188--197,
  Association for Computational Linguistics.

\bibitem{sha_joint_2016}
Lei Sha, Sujian Li, Baobao Chang, and Zhifang Sui,
\newblock ``Joint learning templates and slots for event schema induction,''
\newblock in {\em Proceedings of the 2016 Conference of the North American
  Chapter of the Association for Computational Linguistics: Human Language
  Technologies}. pp. 428--434, Association for Computational Linguistics.

\bibitem{li_prefix-tuning_2021}
Xiang~Lisa Li and Percy Liang,
\newblock ``Prefix-tuning: Optimizing continuous prompts for generation,''
\newblock in {\em Proceedings of the 59th Annual Meeting of the Association for
  Computational Linguistics and the 11th International Joint Conference on
  Natural Language Processing (Volume 1: Long Papers)}. pp. 4582--4597,
  Association for Computational Linguistics.

\bibitem{li_kipt_2022}
Haochen Li, Tong Mo, Hongcheng Fan, Jingkun Wang, Jiaxi Wang, Fuhao Zhang, and
  Weiping Li,
\newblock ``{KiPT}: Knowledge-injected prompt tuning for event detection,''
\newblock in {\em Proceedings of the 29th International Conference on
  Computational Linguistics}. pp. 1943--1952, International Committee on
  Computational Linguistics.

\bibitem{velickovic_graph_2022}
Petar Veli{\v c}kovi{\'c}, Guillem Cucurull, Arantxa Casanova, Adriana Romero,
  Pietro Li{\`o}, and Yoshua Bengio,
\newblock ``Graph attention networks,''
\newblock .

\bibitem{vaswani_attention_2017}
Ashish Vaswani, Noam Shazeer, Niki Parmar, Jakob Uszkoreit, Llion Jones,
  Aidan~N Gomez, {\L}ukasz Kaiser, and Illia Polosukhin,
\newblock ``Attention is all you need,''
\newblock in {\em Advances in Neural Information Processing Systems}. vol.~30,
  Curran Associates, Inc.

\bibitem{zhao_graphsmote_2021}
Tianxiang Zhao, Xiang Zhang, and Suhang Wang,
\newblock ``{GraphSMOTE}: Imbalanced node classification on graphs with graph
  neural networks,''
\newblock in {\em Proceedings of the 14th {ACM} International Conference on Web
  Search and Data Mining}, pp. 833--841.

\bibitem{peng_clustering_2018}
Kai Peng, Victor C.~M. Leung, and Qingjia Huang,
\newblock ``Clustering approach based on mini batch kmeans for intrusion
  detection system over big data,''
\newblock vol. 6, pp. 11897--11906.

\bibitem{hruschka_feature_2005}
E.R. Hruschka and T.F. Covoes,
\newblock ``Feature selection for cluster analysis: an approach based on the
  simplified silhouette criterion,''
\newblock in {\em International Conference on Computational Intelligence for
  Modelling, Control and Automation and International Conference on Intelligent
  Agents, Web Technologies and Internet Commerce ({CIMCA}-{IAWTIC}'06)},
  vol.~1, pp. 32--38.

\bibitem{li_document-level_2021}
Sha Li, Heng Ji, and Jiawei Han,
\newblock ``Document-level event argument extraction by conditional
  generation,''
\newblock in {\em Proceedings of the 2021 Conference of the North American
  Chapter of the Association for Computational Linguistics: Human Language
  Technologies}. pp. 894--908, Association for Computational Linguistics.

\bibitem{lu_text2event_2021}
Yaojie Lu, Hongyu Lin, Jin Xu, Xianpei Han, Jialong Tang, Annan Li, Le~Sun,
  Meng Liao, and Shaoyi Chen,
\newblock ``Text2event: Controllable sequence-to-structure generation for
  end-to-end event extraction,''
\newblock in {\em Proceedings of the 59th Annual Meeting of the Association for
  Computational Linguistics and the 11th International Joint Conference on
  Natural Language Processing (Volume 1: Long Papers)}. pp. 2795--2806,
  Association for Computational Linguistics.

\bibitem{yang_exploring_2019}
Sen Yang, Dawei Feng, Linbo Qiao, Zhigang Kan, and Dongsheng Li,
\newblock ``Exploring pre-trained language models for event extraction and
  generation,''
\newblock in {\em Proceedings of the 57th Annual Meeting of the Association for
  Computational Linguistics}. pp. 5284--5294, Association for Computational
  Linguistics.

\bibitem{nguyen_one_2019}
Trung~Minh Nguyen and Thien~Huu Nguyen,
\newblock ``One for all: Neural joint modeling of entities and events,''
\newblock vol. 33, no. 1, pp. 6851--6858.

\bibitem{lin_joint_2020}
Ying Lin, Heng Ji, Fei Huang, and Lingfei Wu,
\newblock ``A joint neural model for information extraction with global
  features,''
\newblock in {\em Proceedings of the 58th Annual Meeting of the Association for
  Computational Linguistics}. pp. 7999--8009, Association for Computational
  Linguistics.

\bibitem{lin_eliciting_2021}
Jiaju Lin, Jin Jian, and Qin Chen,
\newblock ``Eliciting knowledge from language models for event extraction,''
\newblock .

\bibitem{ferguson_semi-supervised_2018}
James Ferguson, Colin Lockard, Daniel Weld, and Hannaneh Hajishirzi,
\newblock ``Semi-supervised event extraction with paraphrase clusters,''
\newblock in {\em Proceedings of the 2018 Conference of the North American
  Chapter of the Association for Computational Linguistics: Human Language
  Technologies, Volume 2 (Short Papers)}. pp. 359--364, Association for
  Computational Linguistics.

\end{thebibliography}

\end{document}